\documentclass[letterpaper, 10 pt, conference]{ieeeconf}  

\IEEEoverridecommandlockouts                              

\usepackage{amsfonts, dsfont, mathtools, amsmath, amssymb, bbold, mathrsfs, bm, upgreek}
\usepackage{graphicx, float, epsfig, color, psfrag, adjustbox, enumerate}
\usepackage[font=small]{caption}
\usepackage{refcount, url, cleveref}
\usepackage[noadjust]{cite}
\usepackage[usenames,dvipsnames,svgnames,table]{xcolor}

\title{\LARGE \bf
Statics of continuum planar grasping
}

\author{Udit Halder$^{1}$
\thanks{$^{1}$Department of Mechanical Engineering, University of South Florida. 
  Corresponding e-mail:  {\tt\small udithalder@usf.edu}}%
  \thanks{The author is thankful to Dr. Ekaterina Gribkova for insightful discussions on octopus grasping.}
}
\def\R{{\mathds{R}}}

\def\0{{\mathbb{0}}}
\def\1{{\mathds{1}}}

\newcommand{\norm}[1]{\left\lVert#1\right\rVert}

\newcommand{\abs}[1]{\left| #1 \right|}

\definecolor{db}{RGB}{23,20,119}
\definecolor{dg}{RGB}{2,101,15}

\newtheorem{proposition}{Proposition}[section]
\newtheorem{definition}{Definition}
\newtheorem{remark}{Remark}

\usepackage{upgreek}
\newcommand{\dif}{\mathrm{d}}
\newcommand{\transpose}{\intercal}

\newcommand{\material}[1]{
	\ifthenelse{\equal{#1}{\kappa}}{\upkappa}{
	\ifthenelse{\equal{#1}{\nu}}{\upnu}{
	\ifthenelse{\equal{#1}{\omega}}{\upomega}{
	\ifthenelse{\equal{#1}{\sigma}}{\upsigma}{
	\ifthenelse{\equal{#1}{\theta}}{\uptheta}{
	\mathsf{#1}}}}}}
}

\newcommand{\object}{\text{o}}
\newcommand{\tangent}{\text{t}}
\newcommand{\normal}{\text{n}}
\newcommand{\Tangent}{\mathsf{t}}
\newcommand{\Normal}{\mathsf{n}}

\newcommand{\unitx}{\mathsf{e}_1}
\newcommand{\unity}{\mathsf{e}_2}

\newcommand{\wrench}{\mathsf{w}}
\graphicspath{{figures/}}

\begin{document}
\bstctlcite{BSTcontrol} 
\maketitle
\thispagestyle{empty}
\pagestyle{empty}


\begin{abstract}
Continuum robotic grasping, inspired by biological appendages such as octopus arms and elephant trunks, provides a versatile and adaptive approach to object manipulation. Unlike conventional rigid-body grasping, continuum robots leverage distributed compliance and whole-body contact to achieve robust and dexterous grasping. This paper presents a control-theoretic framework for analyzing the statics of continuous contact with a planar object. The governing equations of static equilibrium of the object are formulated as a linear control system, where the distributed contact forces act as control inputs. To optimize the grasping performance, a constrained optimal control problem is posed to minimize contact forces required to achieve a static grasp, with solutions derived using the Pontryagin Maximum Principle. Furthermore, two optimization problems are introduced: (i) for assigning a measure to the quality of a particular grasp, which generalizes a (rigid-body) grasp quality metric in the continuum case, and (ii) for finding the best grasping configuration that maximizes the continuum grasp quality. Several numerical results are also provided to elucidate our methods. 
\end{abstract}



\section{Introduction} \label{sec:intro}

Grasping and manipulation have been fundamental challenges in robotics, traditionally studied in the context of rigid-body systems~\cite{murray1994mathematical,
bicchi2000robotic, li1989grasping}. The study of grasping an object involves many different aspects, ranging from the statics to the dynamics and control of the grasp~\cite{bicchi2000robotic}. In particular, the statics of a grasp describes the equilibrium of the object, considering the contact effects by the grasping agent (e.g., fingers of a robotic hand) and all other external forces and couples (e.g., gravity). It also constrains the allowable contact forces on the object, thus leading to the concepts of grasp quality and planning for the most efficient grasp~\cite{ferrari1992planning, nguyen1988constructing, nguyen1986synthesis}. On the other hand, the study of dynamics and control considers the kinematics and dynamics of the grasping agent -- together with the object being grasped, and control (e.g., joint torques of a robotic manipulator) synthesis for grasping and  manipulation~\cite{li1989grasping, yoshikawa2010multifingered}.

In recent years, the field of soft robotics, especially bioinspired soft robotic systems, has gained attention among roboticists, biologists, and control theorists because soft robots offer remarkable advantages in adaptability, dexterity, and safety due to their intrinsic compliance \cite{rus2015design, laschi2012soft, kim2013soft}. Continuum manipulators, such as those modeled after octopus arms, elephant trunks, and plant vines, provide significant benefits over conventional rigid grippers, particularly in unstructured environments and delicate grasping tasks~\cite{wang2024spirobs, gazzola2018forward}. substantial advancements have been made in the past two decades in designing continuum manipulators and in their various applications, including grasping and manipulation~\cite{mehrkish2021comprehensive, giri2011continuum, li2013autonomous, mehrkish2023multiple, tekinalp2024topology}. 

A key advantage of continuum grasping is that the whole compliant body of the manipulator may be used for grasping, which includes point contacts, distributed contacts, and a combination of the both (see~\cite{mehrkish2021comprehensive} for a classification of different kinds of continuum grasping). Previous model-based studies in continuum soft robotics have largely focused on the theoretical modeling and control strategies for soft robots \cite{della2023model, chang2023energy, wang2022sensory, gazzola2018forward}. Despite these advances, a comprehensive mathematical framework for understanding the grasping and manipulation problem for continuum robots remains relatively less explored~\cite{haibin2018modeling, wu2024bionic}. 

In this paper, we extend the theory of grasping for point contacts~\cite{murray1994mathematical} to the case of distributed contact points, as a step towards constructing a control-theoretic framework for continuum grasping. Here we only consider the planar case for the simplicity of exposition. There are several key aspects of this work. First, we show that the statics of the continuum grasping problem can be expressed by means of a linear `time'\footnote{where an arclength parameter acts as the time variable} varying system. Therefore, the theory of linear systems ~\cite{brockett2015finite, hespanha2018linear} and their reachability properties~\cite{varaiya2000reach, kurzhanski2002reachability} can be applied to study the continuum grasp statics. 

Next, we formulate a series of optimization problems for planning efficient grasps. First, we pose the problem of grasping a given object with minimum contact forces as a constrained optimal control problem whose solution is described by Pontryagin's Maximum Principle~\cite{pontryagin1962mathematical, liberzon2011calculus}. 
This leads to a second-level optimization problem which defines a metric to assess the quality of a grasp. This metric may be regarded as a continuum analog of the similar grasp quality measures in literature~\cite{roa2015grasp, ferrari1992planning, nguyen1988constructing, li1988task}. The last optimization problem then attempts to find the optimal grasping configuration in order to maximize the grasp quality measure. Finally, we provide numerical results demonstrating our methods for a few planar objects, including a circle, an ellipse, and a deformed circle. We note that while this paper is focused on planar grasping, our models and analyses can be extended to the 3D case.

\begin{figure*}[t]
	\centering
	\includegraphics[width=0.75\textwidth, trim = {100pt 110pt 200pt 70pt}]{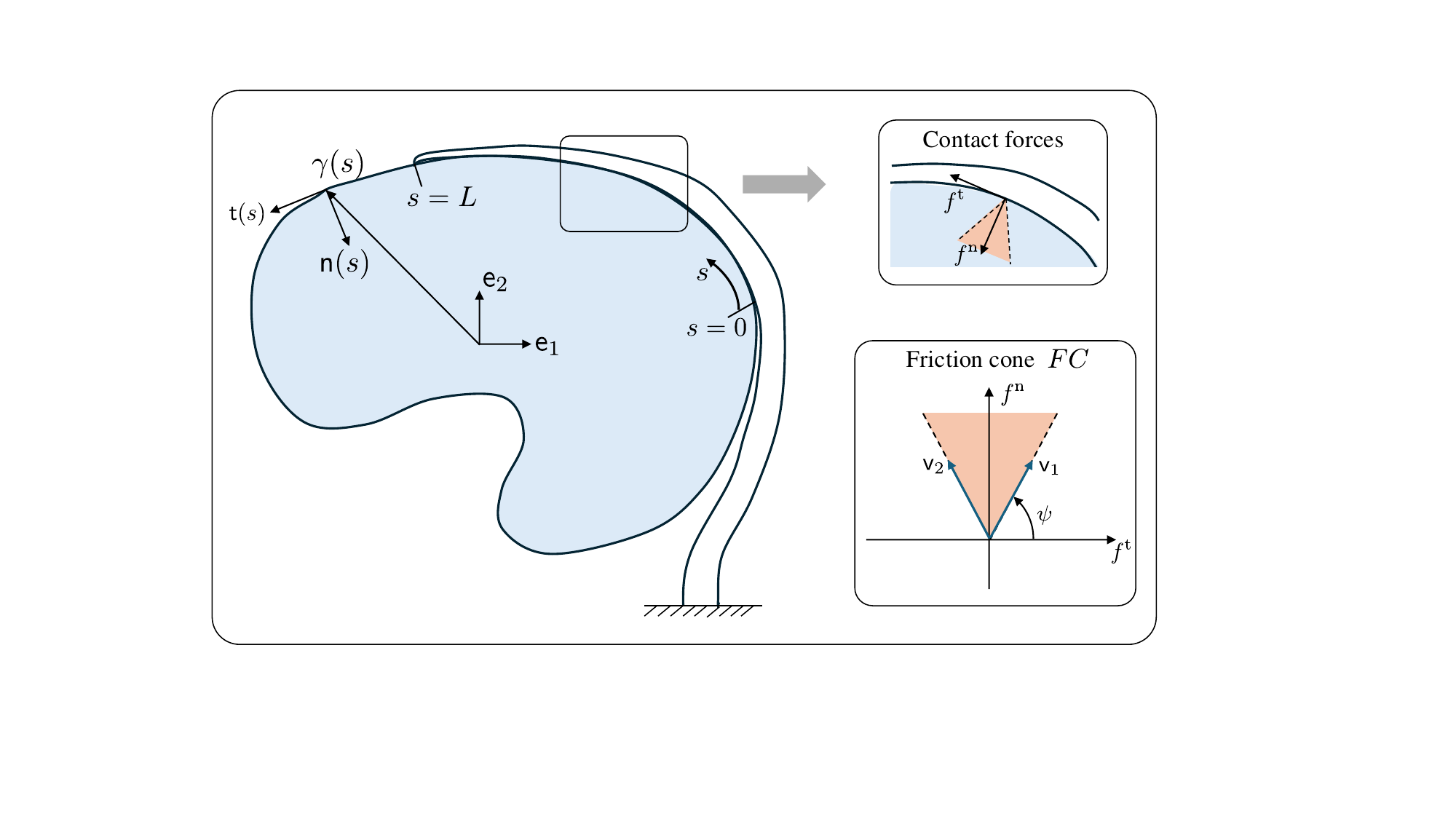}
	\caption{A schematic of continuum grasping of a planar object. The boundary $\gamma$ of the object is parameterized by its arclength $s \in [0, L_\object]$. At each  $s$, there is a moving frame on the boundary $\{\tangent (s), \normal (s) \}$. A soft arm grasps the object by making continuous contact in the interval $[0, L]$. The contact forces per unit length $(f^\tangent (s), f^\normal (s))$ are shown in the top right inset. The constraints of the contact forces due to friction or the friction cone $FC$ is depicted by the orange cone and is elaborated in the bottom right inset.}
	\label{fig:modeling}
	\vspace{-15pt}
\end{figure*}

The remainder of this paper is structured as follows. Section~\ref{sec:modeling} presents the mathematical modeling of the planar object, continuum contact, and the statics of planar continuum grasping. Further analysis of the continuum grasp as a linear time varying system is provided in Sec.~\ref{sec:continuum_grasp_map}. Three optimization problems are formulated in Sec.~\ref{sec:grasp_planning}, which include the continuum analog of grasp quality metric. Numerical results are provided in Sec.~\ref{sec:numerics}, and the paper is concluded with directions of future research in Sec.~\ref{sec:conclusion}.

\section{Modeling} \label{sec:modeling}
In this section, we present models for a planar object, contact forces due to a continuous grasp, and overall statics of the object. 
\subsection{Defining the object}
We consider an object as a planar rigid body with its center of mass at the origin, with the fixed laboratory frame $\{\unitx, \unity \}$ (Fig.~\ref{fig:modeling}). We assume that it has a smooth boundary, i.e. its boundary is represented by a planar smooth curve, parameterized by its arclength $s$, $\gamma\,:\,[0, L_\object] \rightarrow \R^2, s \mapsto (x(s), y(s))$, where $L_\object$ is the length of the boundary, $\gamma(s) = (x(s), y(s))$ is the position of the curve at arclength $s$. Notice that $\gamma$ is a simple closed curve, so that $\gamma (0) = \gamma (L_\object)$.

At each arclength $s$, we attach a moving frame $\{\Tangent(s), \Normal(s)\}$, where $\Tangent(s)$ is the unit tangent vector and $\Normal(s)$ is the unit normal vector so that they define a right-handed orthonormal frame on $\gamma$. Let us denote the angle $\phi(s) \in S^1$ that describes the moving frame, i.e. $\Tangent(s) = \cos \phi \, \unitx + \sin \phi \, \unity$ and $\Normal(s) = -\sin \phi \, \unitx + \cos \phi \, \unity$. Then, the curve $\gamma$ is described by~\cite{bishop1975there}
\begin{align}
\begin{split}
\frac{\dif \gamma (s)}{\dif s}  &= \begin{bmatrix} \cos \phi (s) \\ \sin \phi (s) \end{bmatrix} = \Tangent(s) \\
\frac{\dif \phi(s)}{\dif s} &= \kappa(s)
\end{split}
\label{eq:gamma}
\end{align}
where $\kappa (s)$ is the planar curvature function. 

\subsection{Contact forces} \label{sec:contact_forces}
We are interested in studying the continuum contact of the object with a soft slender body. Without loss of generality, let the contact region be $[0, L]$, where $L \leq L_\object$ is the length of contact. 

At each $s \in [0, L]$, we model the contact as a point contact, and let the contact force per unit length be denoted as $f_c (s) = f^\tangent (s) \, \Tangent (s) + f^\normal (s) \, \Normal(s)$, where $f^\tangent$ and $f^\normal$ are the tangent and normal components of the contact force, respectively (see Fig.~\ref{fig:modeling}). In the laboratory frame, the contact force per unit length can be written in the matrix-vector form as
\begin{align*}
f_c (s) = R (\phi(s)) \begin{bmatrix} f^\tangent (s) \\ f^\normal (s) \end{bmatrix} = R(\phi(s)) f(s)
\end{align*}
where $R(\theta) := \begin{bmatrix*}[r] \cos \theta & - \sin \theta \\ \sin \theta & \cos \theta \end{bmatrix*}$ is the planar rotation matrix, and $f(s) := \begin{bmatrix} f^\tangent (s) \\ f^\normal (s) \end{bmatrix}$ is the contact force per unit length expressed in the local frame.

Moreover, we consider the Coulomb friction model for contact~\cite{murray1994mathematical}, in which the tangent component of the contact force is restricted as follows
\begin{align}
\abs{f^\tangent} \leq \mu f^\normal, ~~ f^\normal \geq 0 
\label{eq:friction_model}
\end{align} 
where $\mu$ is the static coefficient of friction. The condition~\eqref{eq:friction_model} makes sure that there is no slipping. 

In addition to the translational force, the contact force $f_c (s)$ induces a torque $\tau (s)$ on the object about the unit vector coming out of plane
\begin{align*}
\tau  = \gamma \times f_c  = (\gamma^\perp)^\transpose R(\phi) f
\end{align*}  
where $\mathsf{a} \times \mathsf{b} = \text{det} \begin{bmatrix}\mathsf{a} & \mathsf{b} \end{bmatrix}$ is the 2D cross product, and $\gamma^\perp = R(\pi/2) \gamma$ is the perpendicular vector to $\gamma$, obtained by rotating it counterclockwise by an angle of $\pi/2$. 

Taken together, the wrench $\wrench (s) = (f_c (s), \tau (s)) \in \R^3$ acting on the object per unit length of contact is 
\begin{align}
\wrench_c(s) = \begin{bmatrix} I_2 \\ \gamma^\perp (s) \end{bmatrix} R(\phi(s)) f(s) = G(s) f(s) 
\label{eq:wrench_local}
\end{align}
where $I_n$ is the $n\times n$ identity matrix, and the matrix $G(s) := \begin{bmatrix} I_2 \\ \gamma^\perp (s) \end{bmatrix} R(\phi(s))$ maps the local contact forces to the resulting wrench in the object frame of reference. This map $G(s) : \R^2 \rightarrow \R^3$ is known as the \textit{Grasp map} for point contacts~\cite{murray1994mathematical}. 

Finally, the net wrench $w(s)$ acting on the object for continuous grasp in the interval $[0, s]$ is found by integrating~\eqref{eq:wrench_local}
\begin{align}
w (s) = \int_0^s \wrench_c(s) \, \dif s = \int_0^s G(s) f(s) \, \dif s 
\label{eq:wrench_total}
\end{align}
The resultant wrench for the grasp in $[0, L]$ is therefore
\begin{align*}
w_L := w(L) = \int_0^L G(s) f(s) \, \dif s
\end{align*}
Notice that in addition to the local contact forces $f(s)$, the total wrench also depends on the boundary curve $\gamma$ and the length of the grasp $L$.

\smallskip
\begin{remark} 
{\bf (Modeling contact mechanics)} Compliance in soft robotic grasping is of central importance, because the soft robotic agent has the ability to conform to and slide along the object's surface, rather than maintaining a single grasping configuration. However, adequate modeling of the contact mechanics in such scenarios remains a significant challenge~\cite{bicchi2000robotic}, necessitating the use of both analytical and numerical methods.
Theoretical models often draw from elasticity theory and contact mechanics to establish fundamental force-displacement relationships, including the Hertzian model~\cite{johnson1987contact} and tangential compliance models~\cite{popov2019handbook}. 
We defer the incorporation of such models into our framework as future work (see also Remark~\ref{remark:generating_contact_forces}).
\end{remark}


\subsection{Grasp statics}
Suppose there is an external wrench acting on the object $w_e \in \R^3$. The continuum grasp will be in a static equilibrium if
\begin{align}
w_L = - w_{e}
\label{eq:grasp_statics}
\end{align}
i.e., the resultant contact wrench must balance all the external wrench acting on the object. 

\smallskip
\begin{definition}
If for a particular grasp, \eqref{eq:grasp_statics} is satisfied for all $w_{e} \in \R^3$, then we call that grasp \textit{force-closure}\footnote{we use the term force-closure as opposed to wrench-closure, to be consistent with the literature~\cite{murray1994mathematical}}.
\end{definition}

\medskip
In the remainder of the paper, we study the static equation \eqref{eq:grasp_statics}, characterize when might a grasp be force-closure, define the quality of a particular grasp, and plan efficient grasping by designing the local contact forces $f(s)$ and also by choosing optimal grasp configuration. 

\smallskip
\begin{remark} \label{remark:generating_contact_forces}
{\bf (Generating contact forces)} The contact forces $f(s)$ may be generated by proper actuation of the soft manipulator. A body of literature has studied the continuum mechanics and the dynamics of soft and slender manipulators~\cite{della2023model, armanini2023soft}, as well as modeling the effects of their internal actuation (e.g., muscles)~\cite{chang2021controlling, wang2022sensory} using the Cosserat rod theory~\cite{antman1995nonlinear}. In this class of models, a planar soft arm is described by the curve of its centerline, which may be regarded as the Bertrand mate~\cite{zhang2004boundary, bishop1975there} of the boundary curve $\gamma$ of the object. Then, the statics equation \eqref{eq:grasp_statics} can be augmented to also include the equations of statics of the soft arm (see e.g.,~\cite[p. 90]{antman1995nonlinear}). This combined system of equations leads to a relationship between the internal elasticity of the soft arm, its internal actuation (muscle forces), and the contact forces. However, a detailed discussion of this topic goes beyond the scope of the current work.
\end{remark}

\section{The continuum grasp map} \label{sec:continuum_grasp_map}

Let us take the local contact forces $f(s)$ from the set of piecewise continuous functions in the interval $[0, L]$
\begin{align*}
\mathcal{U} = \{ f : [0, L] \rightarrow \R^2, f~ \text{piecewise continuous} \}
\end{align*}
Moreover, the pointwise friction constraint \eqref{eq:friction_model} means that the contact forces must belong to the \textit{friction cone}~$FC :=\{ (f^\tangent, f^\normal)~:~ \abs{f^\tangent} \leq \mu f^\normal, f^\normal \geq 0 \}$, pointwise in $s$. Define the set
\begin{align*}
\mathcal{U}_{FC} := \{ f \in \mathcal{U}~ \text{and } f(s) \in FC, \forall s \in [0, L] \}
\end{align*}
Then, a viable contact force that obeys the friction constraints must belong to $\mathcal{U}_{FC}$.

\begin{remark}
The condition $f^\normal \geq 0$ says that the soft arm can only push the object, but not exert pulling forces ($f^\normal <0$). However, in the context of bioinspired soft robotics, that constraint may be relaxed to allow for a wider range of contact forces. For example, the sucker-mediated grasping in octopus arms uses adhesion forces, along with tangential friction forces~\cite{kier2002structure, tramacere2013morphology}, and a number of attempts have been made to mimic this behavior in robotic systems~\cite{kim2023octopus, xie2020octopus}. 
\end{remark}

\smallskip
We then rewrite \eqref{eq:wrench_total} in its differential form as
\begin{align}
\frac{\dif w}{\dif s}(s) = G(s) f(s), ~~ w(0) = 0
\label{eq:wrench_total_linear} 
\end{align}
Therefore we recognize the resultant wrench equation as a linear time varying system (here, the arclength variable $s$ acts as the `time'), with the local contact forces acting as `controls' and the matrix $G(s)$ is the `input matrix'. Consequently, the grasp statics equation~\eqref{eq:grasp_statics} is essentially the reachability question of the linear system~\eqref{eq:wrench_total_linear}.  

We call the map 
\begin{align}
\begin{split}
\mathcal{G}~&:~ \mathcal{U}_{FC} \rightarrow \R^3 \\
\mathcal{G}\cdot f &= \int_0^L G(s) f(s) \, \dif s
\end{split}
\end{align}
the \textit{grasp map} and this is the continuum analog of the grasp map in discrete contact case~\cite{murray1994mathematical}. The reachability properties of the linear system \eqref{eq:wrench_total_linear} then boils down to studying the reachable set of $\mathcal{G}$. We immediately have the following result. 

\smallskip
\begin{proposition} \label{prop:controllability}
Suppose the contact forces $f \in \mathcal{U}$, i.e., $f^\normal < 0$ is allowed, along with relaxing the friction constraints. Then, the linear control system~\eqref{eq:wrench_total_linear} is controllable. In other words, for any external wrench vector $w_e \in \R^3$, there exists a contact force $f \in \mathcal{U}$, such that \eqref{eq:grasp_statics} is satisfied, i.e. the grasp is force-closure.  
\end{proposition}

\begin{proof}
Indeed, we calculate the controllability gramian
\begin{align}
W &= \int_0^L G(s) G^\transpose (s) \, \dif s \nonumber \\
  &= \int_0^L \begin{bmatrix} I_2 \\ (\gamma^\perp)^\transpose (s) \end{bmatrix} R(\phi(s)) R^\transpose (\phi(s)) \begin{bmatrix} I_2  & \gamma^\perp (s) \end{bmatrix} ~ \dif s \nonumber \\
  &= \int_0^L \begin{bmatrix} I_2 & \gamma^\perp (s) \\ (\gamma^\perp)^\transpose (s) & \norm{\gamma(s)}^2 \end{bmatrix} ~ \dif s \nonumber \\
  &= \begin{bmatrix} L I_2 & \bar{\gamma}^\perp \\ (\bar{\gamma}^\perp)^\transpose & \hat{\gamma}^2  \end{bmatrix} ~ \dif s
  \label{eq:controllability_gramian}
\end{align}
where $\bar{\gamma} = \int_0^L \gamma (s) \, \dif s$ and $\hat{\gamma}^2 = \int_0^L \norm{\gamma}^2 \, \dif s$. We calculate
\begin{align*}
\text{det} (W) = L (L \hat{\gamma}^2 - \norm{\bar{\gamma}}^2 ) 
\end{align*}
Using the Cauchy-Schwarz inequality, it can be readily seen that $\text{det}(W) >0$ for any closed curve $\gamma$ that is described by \eqref{eq:gamma}. Therefore, the controllability grammian is invertible, and thus any wrench vector $w \in \R^3$ may be reached from 0. 
\end{proof}

Proposition~\ref{prop:controllability} is the continuum analog of the condition of the grasp map being surjective. However, we need to consider the effects of friction, in which case the proposition no longer applies. To assess the reachable set of $\mathcal{G}$, we proceed as follows. 

Denote by $\mathsf{v}_1$ and $\mathsf{v}_2$ the two unit vectors in the direction of the `extreme rays' of the cone $FC$ (see Fig.~\ref{fig:modeling})
\begin{align*}
\mathsf{v}_1 = \begin{bmatrix} \cos \psi \\ \sin \psi \end{bmatrix}, \quad \mathsf{v}_2 = \begin{bmatrix} -\cos \psi \\ \sin \psi \end{bmatrix}
\end{align*} 
where $\psi = \tan^{-1} (1/\mu)$. Then, any contact force $f(s)$ can be expressed as 
\begin{align}
f(s) = u_1(s) \mathsf{v}_1 + u_2(s) \mathsf{v}_2, ~~ u_1(s), u_2(s) \geq 0~ \forall s
\label{eq:contact_forces_cone}
\end{align}
In other words, $f(s)$ is in the conical hull of $\{\mathsf{v}_1, \mathsf{v}_2 \}$. We can then represent the reachable set, i.e. all the wrenches that can be reached from 0 as
\begin{align}
w(L) = \left(\int_0^L u_1(s) G(s) \, \dif s \right) \mathsf{v}_1 + \left(\int_0^L u_2(s) G(s) \, \dif s \right) \mathsf{v}_2 
\end{align}
where $u_1(s), u_2(s) \geq 0$ for all $s$. Since $G$ depends on the curve $\gamma$, the reachable set can be approximated numerically~\cite{varaiya2000reach, hwang2005polytopic, kurzhanski2002reachability}.

\smallskip
\begin{remark} ({\bf Constructing different kinds of grasps})
Considering the contact forces $f(s)$ as piecewise continuous functions allows us to model a variety of contact scenarios. For example, instead of continuum contact on $[0, L]$, if we only have contacts at $n$ discrete points $0\leq s_k \leq L, k= 1, 2, \cdots, n$, then $f(s) = f_k \delta (s-s_k)$, where $f_k \in FC$ is the contact force at the $k$-th contact, and $\delta(\cdot)$ is the delta function. Total wrench in this case is
\begin{align*}
w_c = \int_0^L G(s) f(s) = \sum_{k=1}^n G(s_k) f_k  
\end{align*} 
Consequently the grasp map becomes
\begin{align*}
\mathcal{G} = \begin{bmatrix} G(s_1) & G(s_2) & \cdots & G(s_n) \end{bmatrix} \in \R^{3\times 2n}.
\end{align*} 
This is essentially the grasp map in the theory of grasping for discrete point contacts~\cite{murray1994mathematical}. Similarly, one can construct other types of grasp (e.g., a combination of continuum and discrete contacts) as outlined in~\cite{mehrkish2021comprehensive}.
\end{remark}

\section{Grasp planning} \label{sec:grasp_planning}

In this section, we formulate a series of optimization problems for synthesizing the optimal grasp. 
At the outset, we rewrite equations \eqref{eq:wrench_total_linear}, \eqref{eq:contact_forces_cone} as the following linear control system
\begin{align}
\frac{\dif w}{\dif s}(s) = B(s) u(s), ~~ w(0) = 0
\label{eq:wrench_total_linear_positive} 
\end{align}
where $B(s) = G(s)V = G(s) \begin{bmatrix} \mathsf{v}_1 &\mathsf{v}_2 \end{bmatrix}$ is the input matrix, and $u(s) = \begin{bmatrix} u_1(s) \\ u_2(s) \end{bmatrix}$ is the piecewise-continuous control input such that $u_1(s), u_2(s) \geq 0, \forall s$. 

\subsection{Minimum force grasping} \label{sec:min_force_grasping}
First, for a given wrench $w_e \in \R^3$, a natural LQR problem may be formulated for the linear control system~\eqref{eq:wrench_total_linear_positive} which minimizes the norm of the control input $\norm{u(s)}^2$, such that $w(L) = -w_e$. This problem attempts to find the minimal control effort that can resist a given external wrench $w_e$. Since the controls are constrained, the standard result for linear systems (involving the Riccati equation)~\cite{liberzon2011calculus} does not immediately apply. Accounting also for the ease of implementation, we consider the following free endpoint optimal control problem  
\begin{align}
\begin{split}
\underset{u(\cdot)}{\text{minimize}} ~~ &J(u; w_e) = \frac{1}{2}\int_0^L \norm{u(s)}^2 \, \dif s + \frac{\chi}{2} \norm{w(L) + w_e}^2 \\ 
\text{subject to} \quad &\frac{\dif w}{\dif s} = B(s) u(s), ~ w(0) = 0, w(L)~ \text{free} \\
\text{and} \quad &u_i(s) \geq 0,~~ i = 1, 2, \forall s
\end{split}
\label{eq:optimal_control_problem_1}
\end{align}
Here, $\chi >0$ is a large positive parameter penalizing the deviation of $w(L)$ from $-w_e$. 

Solution to \eqref{eq:optimal_control_problem_1} is readily obtained by the application of the Pontryagin Maximum Principle~\cite{pontryagin1962mathematical, liberzon2011calculus}. By denoting the costate vector $p \in \R^3$, the control Hamiltonian is written as
\begin{align*}
H(w, p, u) = p^\transpose Bu - \frac{1}{2}\norm{u}^2 - \frac{\chi}{2} \norm{w(L) + w_e}^2
\end{align*}
The optimal control is found by maximizing the control Hamiltonian 
\begin{align}
u_i(s) = \max \{ (B(s)^\transpose p(s))_i, 0 \}, ~~ i = 1, 2
\label{eq:optimal_control}
\end{align}
Moreover, the costate $p$ is a constant vector, since it evolves according to the Hamilton's equation
\begin{align*}
\frac{\dif p}{\dif s} = -\frac{\partial H}{\partial w} = 0, 
\end{align*}
along with the transversality condition
\begin{align*}
p(L) = - \frac{\chi}{2} \frac{\partial}{\partial w} \norm{w(L) + w_e}^2 = - \chi (w(L) + w_e)
\end{align*}

Let the optimal solution of \eqref{eq:optimal_control_problem_1} be denoted by $u^*(\cdot)$. For a given $w_e$, denote the optimal cost as
\begin{align*}
J^*(w_e) := J(u^*; w_e)
\end{align*}
This quantity measures the minimum amount of grasp effort needed to resist a given external wrench $w_e$. One can characterize $J^* (w_e)$ under certain assumptions, which is described by the following proposition.

\begin{proposition} \label{prop:J_star}
Suppose there exists a wrench $w_e \in \R^3$, such that the solution of \eqref{eq:optimal_control_problem_1} lies in the interior of the allowable control set, i.e. $u_i(s) >0, \forall s, i = 1, 2$. Then, 
\begin{align}
\begin{split}
J^*(w_e) =& \frac{1}{2}\chi w_e^\transpose (I_3 + \chi \hat{W})^{-1} w_e
\end{split}
\end{align}
where $\hat{W}$ is the controllability gramian of the linear system~\eqref{eq:wrench_total_linear_positive}, $\hat{W} = \int_0^L B(s)B^\transpose (s) \, \dif s$.
\end{proposition}

\begin{proof}
According to the assumption, $u^* (s) = B^\transpose (s) p$ from \eqref{eq:optimal_control}, where $p$ is a constant vector
\begin{align}
p = -\chi (w_L + w_e)
\label{eq:p}
\end{align}
We calculate
\begin{align*}
w_L &= \int_0^L B(s) B^\transpose (s) p \, \dif s 
	= \left( \int_0^L B(s) B^\transpose (s) \, \dif s \right) p \\
	&= \hat{W}p
\end{align*}
Plugging this in \eqref{eq:p}, we get
\begin{align*}
p &= - \chi (I_3 + \chi \hat{W})^{-1} w_e
\end{align*}
where we have used that $\hat{W} \succeq 0$ since $\hat{W}$ is a controllability gramian, and hence $(I_3 + \chi \hat{W})$ is invertible for any $\chi > 0$.

Finally, we find
\begin{align*}
J^*(w_e) &= \frac{1}{2} \int_0^L {u^*}^\transpose (s) u^* (s) \, \dif s + \frac{\chi}{2} \norm{w_L + w_e}^2 \\
		&= \frac{1}{2} p^\transpose \left( \int_0^L B (s) B^\transpose(s) \, \dif s \right) p + \frac{\chi}{2}\cdot \frac{1}{\chi^2} p^\transpose p \\
		&= \frac{1}{2} p^\transpose \hat{W} p + \frac{1}{2\chi} p^\transpose p 
		= \frac{1}{2\chi} p^\transpose \left( I_3 + \chi \hat{W} \right) p \\
		&= \frac{\chi}{2} w_3^\transpose (I_3 + \chi \hat{W})^{-1} w_e
\end{align*}

\end{proof}

\smallskip
\begin{remark}
Consider the case in Proposition~\ref{prop:J_star}. We notice that 
\begin{align*}
\lim\limits_{\chi \rightarrow \infty} w_L = \lim\limits_{\chi \rightarrow \infty} - \chi \hat{W} (I_3 + \chi \hat{W})^{-1}  w_e = - w_e
\end{align*}
Therefore, choosing a large value of $\chi$ yields a desired solution in practice. 
\end{remark}

\smallskip
\begin{remark}
The underlying assumption for the problem \eqref{eq:optimal_control_problem_1} is that $-w_e$ lies in the reachable set of the control system \eqref{eq:wrench_total_linear_positive}. Nevertheless, even if $-w_e$ cannot be reached, the optimal control problem~\eqref{eq:optimal_control_problem_1} may be solved to produce minimum-norm contact forces that best resists $w_e$. However, one might need additional constraints on the maximum applicable contact forces for a meaningful solution.
\end{remark}

\begin{figure*}[t]
	\centering
	\hspace*{-10pt}
	\includegraphics[width=\textwidth, trim = {50pt 110pt 30pt 25pt}, clip = true]{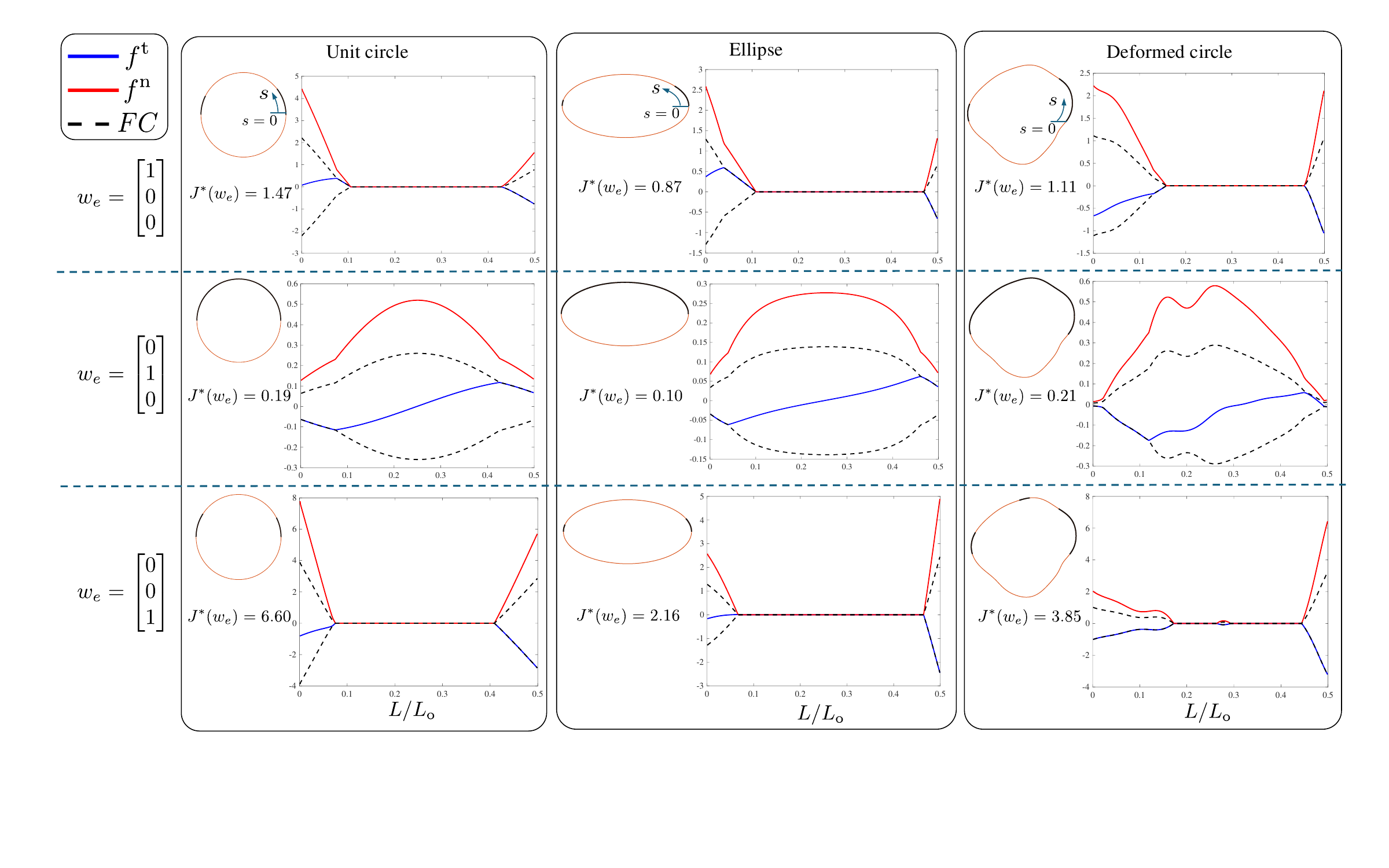}
	\caption{Minimum force grasping as a solution of problem \eqref{eq:optimal_control_problem_1}. For each of the objects (unit circle, ellipse, deformed circle), each graph represents the minimum contact forces required to resist the corresponding external wrench $w_e$. The tangent forces are shown in solid blue lines, the normal forces are shown in solid red lines, and the friction constraints are shown by dashed black lines. The objects are also shown in orange, and the optimal region to be grasped is depicted by solid black lines overlaid on the boundaries. The optimal grasp effort $J^*(w_e)$ is also indicated for each case. 
	}
	\label{fig:min_force}
	\vspace{-15pt}
\end{figure*}

\subsection{Grasp quality}
The concept of grasp quality is well-developed in the literature of grasping~\cite{ferrari1992planning, li1988task, roa2015grasp, rubert2018characterisation}, where a metric is assigned to quantify how `good' a particular grasp is. There have been a few different such quality metrics that have been proposed, and the choice of a particular metric depends on the application under consideration. In this paper, we generalize the Ferrari-Canny grasp quality measure~\cite{ferrari1992planning} for continuum grasping. The Ferrari-Canny metric captures the smallest external wrench that the grasp can resist in any direction, and is defined as the solution to a min-max problem.  

As we have seen in Sec.~\ref{sec:min_force_grasping}, $J^*(w_e)$ is the minimum grasp effort needed to resist a particular $w_e$. Therefore, a lower value of $J^*(w_e)$ indicates that the wrench $w_e$ can be more efficiently resisted, and thus the `quality' of that grasp is `good', i.e. $J^*$ is inversely proportional to the quality of a grasp. Since we do not have control over external wrenches, following~\cite{ferrari1992planning}, we introduce the grasp quality measure for a continuum grasp by considering the worst case scenario
\begin{align}
\mathcal{Q} = \frac{1}{\tilde{J}^*}, ~\text{where} ~ \tilde{J}^* := \underset{w_e, ~\norm{w_e}=1}{\max} {J^*(w_e)}
\label{eq:grasp_quality}
\end{align}
The grasp quality metric $\mathcal{Q}$ is therefore the worst, over all possible wrench directions, of all the most efficient (minimum-force) grasps. A solution to~\eqref{eq:grasp_quality} also gives the direction of the wrench vector in which the optimal minimum-force grasp is the least good. The higher the $\mathcal{Q}$ value is, the better the grasp is.  

We note here that other types of grasp quality metrics (e.g., the smallest singular value of the grasp map metric~\cite{li1988task}) may also be defined that could be more suitable for respective applications.


\subsection{Maximizing grasp quality} \label{sec:maximize_grasp_quality}
Finally, we are ready to formulate the last optimization problem for efficient grasp planning. Notice that the quantities (i) minimum grasp effort $J^*$ and (ii) the grasp quality $\mathcal{Q}$ are both defined for a particular grasp, i.e. for a grasp on the interval $[0, L]$. However, from the grasping agent's (e.g., a soft arm) point of view, this interval does not need to be fixed. By introducing another parameter $s_0 \in [0, L_\object]$, we redefine the grasping interval as $[s_0, (s_0 + L) \, (\text{mod} \, L_\object)]$, where we use the periodicity of the curve $\gamma$. Therefore we have two free parameters -- (i) $s_0$: the arclength at which the grasp starts, and (ii) $L$: the length of the grasp; and thus we write $\mathcal{Q} = \mathcal{Q}(s_0, L)$. 

\begin{figure}[t]
	\centering
	\hspace*{-10pt}
	\includegraphics[width=0.4\textwidth, trim = {120pt 25pt 140pt 25pt}, clip = true]{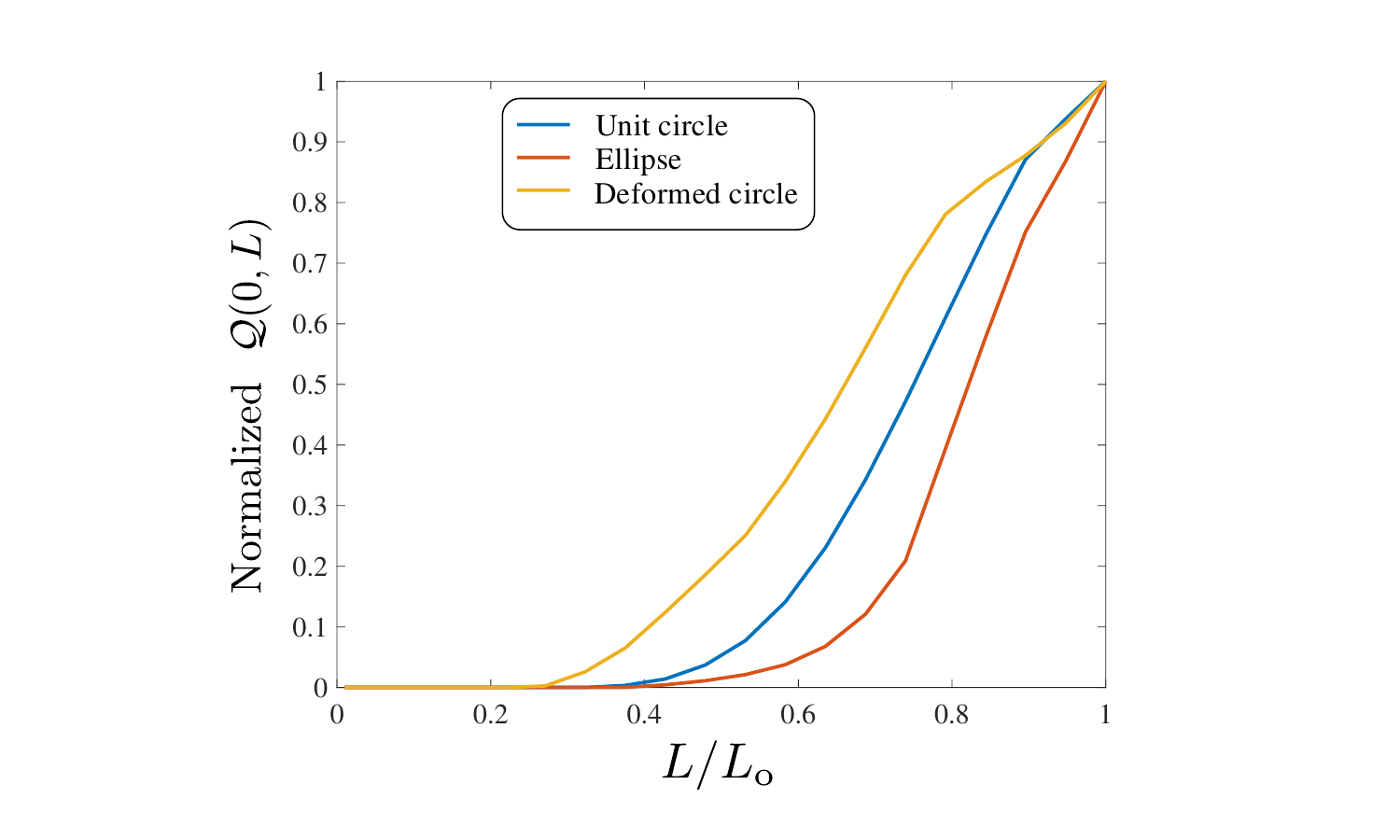}
	\caption{Normalized grasp quality $\mathcal{Q}(0, L)$ as a function of the grasp length $L$, for each of the objects. The grasp quality increases as the grasp length increases.}
	\label{fig:Q_L}
	\vspace{-15pt}
\end{figure}

However, in many practical examples, intuitively it makes sense to increase the length of the grasp in order to improve the grasp quality, i.e. $\mathcal{Q}(s_0, L)$ is monotonously increasing in $L$. This is also verified numerically in Sec.~\ref{sec:numerics_max_grasp_quality} (see Fig.~\ref{fig:Q_L}). Moreover, the length of a soft manipulator may not be flexible. Therefore, for a given $L \in (0, L_\object]$, we solve the following optimization problem
\begin{align}
\underset{s_0 \in [0, L_\object]}{ \text{maximize}} ~~ \mathcal{Q} (s_0; L)
\label{eq:maximize_grasp_quality}
\end{align}
A solution to the optimization problem~\eqref{eq:maximize_grasp_quality} provides the optimal arclength of the boundary where the grasp should be started to achieve the best quality grasp of the object.

\section{Numerical examples} \label{sec:numerics}
In this section, we demonstrate the solutions to the three optimization problems formulated in Sec.~\ref{sec:grasp_planning} for three boundary curves: (i) a circle of radius $r$, (ii) an ellipse with semi-major and semi-minor axes $a$ and $b$, respectively, and (iii) a randomly `deformed' circle (see e.g., the third column of Fig.~\ref{fig:min_force}). For the third boundary, a unit circle is first taken and (pseudo) random sinusoidal oscillatory disturbances are added to obtain a deformed yet smooth boundary curve. For consistency, the seed of the random number generator is kept constant across experiments so that a fixed deformed circle can be generated. Each of the optimization problems are then solved for each boundary. The values of the fixed parameters are: $r = 1, a = 2, b = 1$, friction coefficient $\mu = 0.5$, and $\chi = 5 \times 10^4$. The rest of the parameters are varied and are described in their respective experiments.

\begin{figure*}[t]
	\centering
	\hspace*{-10pt}
	\includegraphics[width=\textwidth, trim = {10pt 55pt 10pt 20pt}, clip = true]{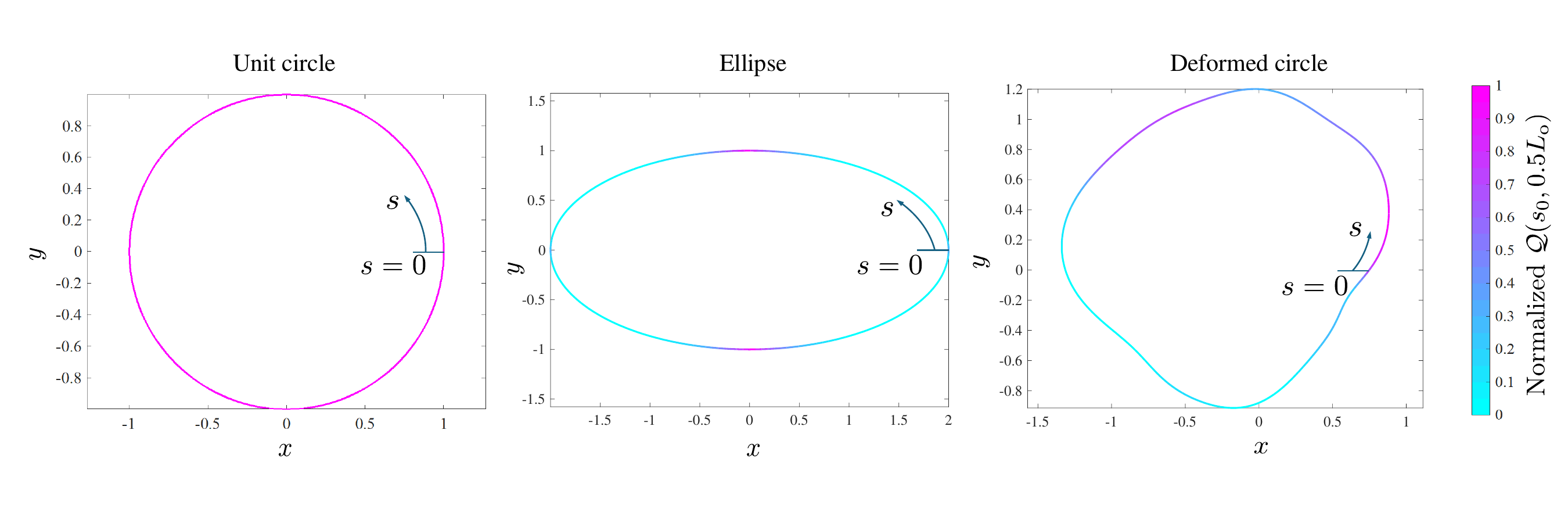}
	\caption{Maximizing grasp quality by solving optimization problem~\eqref{eq:maximize_grasp_quality}. For a fixed $L = 0.5L_\object$, normalized grasp quality $\mathcal{Q}(s_0, L)$ is first calculated as a function of $s_0 \in [0, L_\object]$. These values are then overlaid on top of each of the objects, with the colorbar shown on the right. Therefore, the more pink a point is on the object, the better is the grasp quality if the grasp were to be started at the point. Every point on the circle is equivalent in terms of grasp quality, whereas it is not true for the other two objects.
	}
	\label{fig:optimal_Q}
	\vspace{-15pt}
\end{figure*}

\subsection{Minimum force grasping}
First, the optimal control problem~\eqref{eq:optimal_control_problem_1} is numerically solved to find the optimal contact forces required to resist a given external wrench $w_e$. An iterative approach called the forward-backward algorithm~\cite{chang2020energy, chang2023energy} is used to solve the optimal control problem. In particular, let $w^{(k)}, p^{(k)}$, and $u^{(k)}$ denote the state, costate, and control at the $k$-th iteration. Then, the update rule for the control is given as follows
\begin{align}
\begin{split}
u^{(k+1)} &= \max \left\lbrace u^{(k)} + \eta \frac{\partial H}{\partial u} \left( w^{(k)}, p^{(k)}, u^{(k)} \right), 0 \right\rbrace \\
		 &=  \max \left\lbrace u^{(k)} + \eta \left( B^\transpose p^{(k)} - u^{(k)} \right), 0 \right\rbrace
\end{split}
\label{eq:u_update}
\end{align}
where $\eta > 0$ is a small step-size parameter. The $\max$ operator in the $u$ update rule~\eqref{eq:u_update} makes sure the control constraints are satisfied at every iteration. For each of the objects, the following parameters are used: 
length of grasp $L = 0.5L_\object$, where $L_\object$ is the length of the object, $\eta = 10^{-6}$; and the external wrench is taken from the set $w_e \in \left\lbrace \begin{bmatrix}1 & 0 & 0\end{bmatrix}^\transpose, \begin{bmatrix}0 & 1 & 0\end{bmatrix}^\transpose, \begin{bmatrix}0 & 0 & 1\end{bmatrix}^\transpose \right\rbrace$. Once the optimal controls $u$ are found, the optimal contact force vector is calculated by $f = Vu$.

The results are shown in Fig.~\ref{fig:min_force}. The normal components ($f^\normal$) of the contact force vector are shown using solid blue lines and the tangential components ($f^\tangent$) are shown in solid red lines. The friction constraints ($\abs{f^\tangent} \leq \mu f^\normal$) are also indicated by dashed black lines. It is observed that the optimal grasp does not necessarily need to be truly continuum. Consequently, the portions of the boundary curves required to grasp by the optimal control (i.e. where $f^\normal >0$) are shown by solid black lines overlaid on the boundary curves. As is seen from Fig.~\ref{fig:min_force}, except for $w_e = \begin{bmatrix} 0 & 1 & 0\end{bmatrix}^\transpose$, it is optimal to grasp only the beginning and the end parts of the grasping interval $[0, 0.5L_\object]$. This kind of grasping is characterized by~\cite{mehrkish2021comprehensive}. On the other hand, for $w_e = \begin{bmatrix} 0 & 1 & 0\end{bmatrix}^\transpose$, truly continuum grasping is observed for each of the boundaries. Finally, a three part grasp is found for the deformed circle case for $w_e = \begin{bmatrix} 0 & 0 & 1\end{bmatrix}^\transpose$.

The optimal grasp effort $J^*(w_e)$ is also computed for each case and is reported on Fig.~\ref{fig:min_force}. It is observed that pure force in the $y$ direction is most easily resisted for each of the objects, while the pure couple is the hardest to resist. On the other hand, the grasp effort is the lowest for the ellipse among the objects, for each $w_e$. It is to be noted that these conclusions are the consequences of the chosen grasp region $[0, 0.5L_\object]$, and different results might ensue if the grasp interval is varied.

\subsection{Continuum grasp quality}
Next, we calculate the continuum grasp quality for each of the boundaries by solving the optimization problem~\eqref{eq:grasp_quality}. For this experiment, we only consider grasp in the interval $[0, 0.5L_\object]$. Notice that the constrained maximization problem~\eqref{eq:grasp_quality} can be converted into an unconstrained  problem by writing $w_e$ in the spherical coordinates $(\cos \alpha, \sin \alpha \cos \beta, \sin \alpha \sin \beta)$, and maximizing over $(\alpha, \beta)$. We therefore use Matlab function \textit{fminunc} to minimize $-J^*(w_e)$ (thus maximizing $J^*$). We calculate the grasp quality metric $\mathcal{Q}$ to be $0.0736, 0.0838$, and  $0.1786$, for the three boundary curves, respectively. The worst wrench directions are $\begin{bmatrix} 0.2744 & -0.7087 & 0.6500 \end{bmatrix}^\transpose$ for the circle,    $\begin{bmatrix} 0.1838 & -0.9158 & 0.3572 \end{bmatrix}^\transpose$ for the ellipse, and  $\begin{bmatrix} 0.3640 & -0.4971 & 0.7877 \end{bmatrix}^\transpose$ for the deformed circle. This indicates the deformed circle has the better grasp quality than the other two objects in the interval $[0, 0.5L_\object]$.

\subsection{Maximizing grasp quality} \label{sec:numerics_max_grasp_quality}
Finally, the third level optimization problem~\eqref{eq:maximize_grasp_quality} is solved to find the optimal grasping location for a given object boundary. Here, the  arclength $s_0$ and the length of the grasp $L$ both can be varied. First, by setting $s_0 = 0$, twenty values of $L$ are selected from a uniform discretization of the interval $[0.01 L_\object, L_\object]$, and corresponding $\mathcal{Q} (0, L)$ values (normalized) are plotted in Fig.~\ref{fig:Q_L}. The grasp quality values are seen to be monotonically increasing in $L$ for each boundary. Thus, as described in Sec.~\ref{sec:maximize_grasp_quality}, maximum grasp quality is only calculated for a fixed $L$. Note that the maximization problem~\eqref{eq:maximize_grasp_quality} is nested twice, i.e., it needs to solve \eqref{eq:grasp_quality} in each function evaluation, which in turn needs to solve~\eqref{eq:optimal_control_problem_1} in each of its own function evaluations. Therefore, a discrete search approach is adopted instead of traditional optimization algorithms for feasible computation time. With $L = 0.5L_\object$, the interval $[0, L_\object]$ is uniformly discretized into 200 points for $s_0$, corresponding $\mathcal{Q}(s_0, 0.5L_\object)$ values are calculated, and the maximum is selected from this set.


The results are shown in Fig.~\ref{fig:optimal_Q} by overlaying the normalized $\mathcal{Q}$ values on top of the boundaries itself. It is seen that all points on the circle yield the same grasp quality due to its rotational symmetry. However, the ellipse and the deformed circle exhibit more interesting behavior. It is found that the grasp that maximizes its continuum grasp quality should start at $s_0 = 0.25L_\object$ or $s_0 = 0.75L_\object$ for the ellipse. 
Finally, the optimal $s_0$ is found to be $0.045L_\object$ for the deformed circle. 

\section{Conclusion and future work} \label{sec:conclusion}

This paper presents a control-oriented framework for analyzing the statics of planar continuum grasping, drawing inspiration from soft robotic manipulators and bioinspired grasping mechanisms. Our models generalize the theory of grasping for point contacts to continuous contact points. The equations of static equilibrium of an object are represented in terms of a linear control system, where the control inputs are recognized as the contact forces. Next, the problem of minimum force grasping is posed as an optimal control problem, whose solution is also discussed. A notion of continuum grasp quality is developed, extending a popular grasp quality measure for point contacts. The problem of optimal grasping configuration is also discussed. Several numerical simulations are provided to demonstrate our framework. 

Future work will incorporate the mechanics of a soft arm with the object being grasped, which also captures the contact compliance -- a subject of critical importance in continuum robotic grasping. The kinematics of the relative motion of these two bodies, as well as the dynamics of the combined system will be another direction of research. Furthermore, designing reliable control strategies, especially feedback control~\cite{wang2022sensory, wang2024neural} for continuum robots for adaptively grasping an object based on tactile feedback will also be considered. 

On the other hand, experimental validation of the theoretical predictions using soft robotic prototypes will enhance the applicability of our models to real-world grasping tasks. The insights gained from this work contribute to the broader field of soft robotics and bioinspired manipulation, paving the way for more adaptable and dexterous robotic grasping systems.

\vspace*{-10pt}

\bibliographystyle{IEEEtran}
\bibliography{bibfiles/octopus_papers,bibfiles/reference}

\appendices
\renewcommand{\thelemma}{A-\arabic{section}.\arabic{lemma}}
\renewcommand{\thetheorem}{A-\arabic{section}.\arabic{theorem}}
\renewcommand{\theequation}{A-\arabic{equation}}
\renewcommand{\thedefinition}{A-\arabic{definition}}
\setcounter{lemma}{0}
\setcounter{theorem}{0}
\setcounter{equation}{0}


\end{document}